\documentclass[10pt,a4paper]{article}
\usepackage{lrec2026}
\usepackage{graphicx}
\usepackage{amsmath}
\usepackage{float}
\usepackage{booktabs}
\usepackage{fontspec}
\usepackage{polyglossia}
\setmainlanguage{english}
\setotherlanguage{arabic}
\newfontfamily\arabicfont[Script=Arabic]{Amiri-Regular.ttf}

\title{LLM-Based Financial Sentiment Analysis in Arabic: Evidence from Saudi Markets}

\name{Mona H. Albaqawi$^{1}$, Eman M. Albalkhi$^{2}$, Joud A. Albaiti$^{3}$, Enrico Lopedoto$^{4}$}

\address{
$^{1}$ George Mason University,
$^{2}$ Damascus University \\
$^{3}$ University of Jeddah ,
$^{4}$ City, St George's, University of London \\
\texttt{albaqawih@gmail.com, emanmousaalbalkhi@gmail.com} \\ \texttt{joudalbaiti.7@gmail.com, enrico.lopedoto@city.ac.uk}
}

\abstract{
Investor sentiment shapes financial markets, yet modeling sentiment in Arabic financial contexts remains challenging due to linguistic complexity and limited resources. We present an Arabic NLP framework for large-scale financial sentiment analysis tailored to the Saudi market, integrating official financial news and social media to capture institutional and public investor sentiment. The framework constructs a large Arabic financial corpus through a multi-stage pipeline encompassing data collection, cleaning, deduplication, entity linking, and sentiment annotation. Transformer-based NER combined with a curated company lexicon links textual mentions to canonical company identifiers, with sentiment labels assigned using a five-class scheme. The resulting dataset of 84K samples supports company-level sentiment aggregation and analysis of sentiment dynamics relative to stock market behavior on the Saudi Exchange. Experimental results demonstrate reliable and scalable Arabic financial sentiment analysis.
 \\ \newline }

\begin{document}

\maketitleabstract


\section{Introduction}
Financial news and investor sentiment shape market psychology and influence stock price movements. Prior research demonstrates that sentiment extracted from financial texts associates closely with market behavior, including price fluctuations and volatility \citep{Barberis1997, Ahmad2023}. In Saudi Arabia, where information disseminates rapidly across traditional and digital platforms, quantifying financial content tone provides valuable analytical insights consistent with the sentiment framework of Barberis, Shleifer, and Vishny ~\citep{Barberis1997}, which shows that market prices may deviate from fundamental values due to sentiment-driven trading. Despite growing interest, most Arabic financial sentiment studies rely on traditional machine learning with extensive feature engineering \citep{Ahmad2023}, often failing to capture long-range context and nuanced sentiment.

To address these limitations, this study presents an Arabic NLP framework enabling (i) temporal analysis through near real-time processing, (ii) domain-specific sentiment classification tailored to Arabic financial linguistic patterns, and (iii) analysis of relationships between sentiment dynamics and equity market behavior on the Saudi Exchange. Unlike conventional approaches, the framework adopts large language models for contextual reasoning and semantic robustness \citep{Gu2024, Wang2023}, processing heterogeneous financial text through a unified multi-stage pipeline that reduces noise, mitigates redundancy, and preserves financial semantics. LLMs perform summarization and five-class sentiment labeling, while transformer-based models handle named entity recognition and entity linking. A multi-model agreement strategy enhances labeling reliability and mitigates single-model bias.

Overall, this work contributes a scalable approach to Arabic financial sentiment analysis and demonstrates the applicability of LLM-based pipelines in Arabic NLP research.



\section{Related Work}

Financial sentiment analysis captures qualitative signals embedded in financial texts and links them to market behavior. Early foundational work by Barberis et al. ~\citep{Barberis1997} demonstrated that investor sentiment can drive deviations of market prices from fundamental values. Sentiment analysis techniques have evolved from lexicon-based approaches to machine learning and deep learning methods. Domain-specific sentiment lexicons show effectiveness in capturing polarity in financial discourse \citep{Sohangir2018}, while transformer-based models have advanced financial sentiment classification through contextual representations, as demonstrated by FinBERT \citep{Araci2019}. In Arabic financial contexts, the BORSAH corpus \citep{Alshahrani2018} highlights integrating official financial news with social media to capture investor sentiment dynamics in the Saudi market. Compared to BORSAH (Alshahrani et al., 2018), which focuses on manually annotated Twitter data, our dataset differs primarily in scope and methodology. While BORSAH relies exclusively on Twitter and three-class manual annotation, our corpus integrates both financial news and social media sources and employs a scalable multi-model LLM consensus strategy with a five-class sentiment taxonomy, enabling broader coverage and finer-grained sentiment analysis.

Arabic NLP presents unique challenges due to rich morphology, orthographic variation, and absence of capitalization. Transformer-based approaches are well suited for Arabic NER, achieving robust performance \citep{ElBazi2019}. Arabic language models such as AraBERT \citep{Antoun2020} and CAMeLBERT \citep{Inoue2021} provide contextual embeddings optimized for Arabic morphology. Large-scale Arabic-centric foundation models have emerged, including Jais \citep{Sengupta2023}, AceGPT \citep{Huang2024}, and ALLaM \citep{Abdelali2024}, applied to financial text processing tasks. While these models offer strong abstractive capabilities, concerns regarding hallucination remain prominent. Evaluation frameworks such as AraHalluEval provide methods for assessing hallucination in Arabic language models \citep{Alansari2025}, and recent surveys examine large language models as evaluators \citep{Du2024, Gu2024}.

Overall, existing literature highlights the importance of domain-aware preprocessing, robust entity linking, and careful evaluation when applying NLP techniques to financial sentiment analysis, particularly in Arabic-speaking markets.


\section{System Architecture}
\begin{figure}[t]
    \centering
    \includegraphics[width=0.50\textwidth]{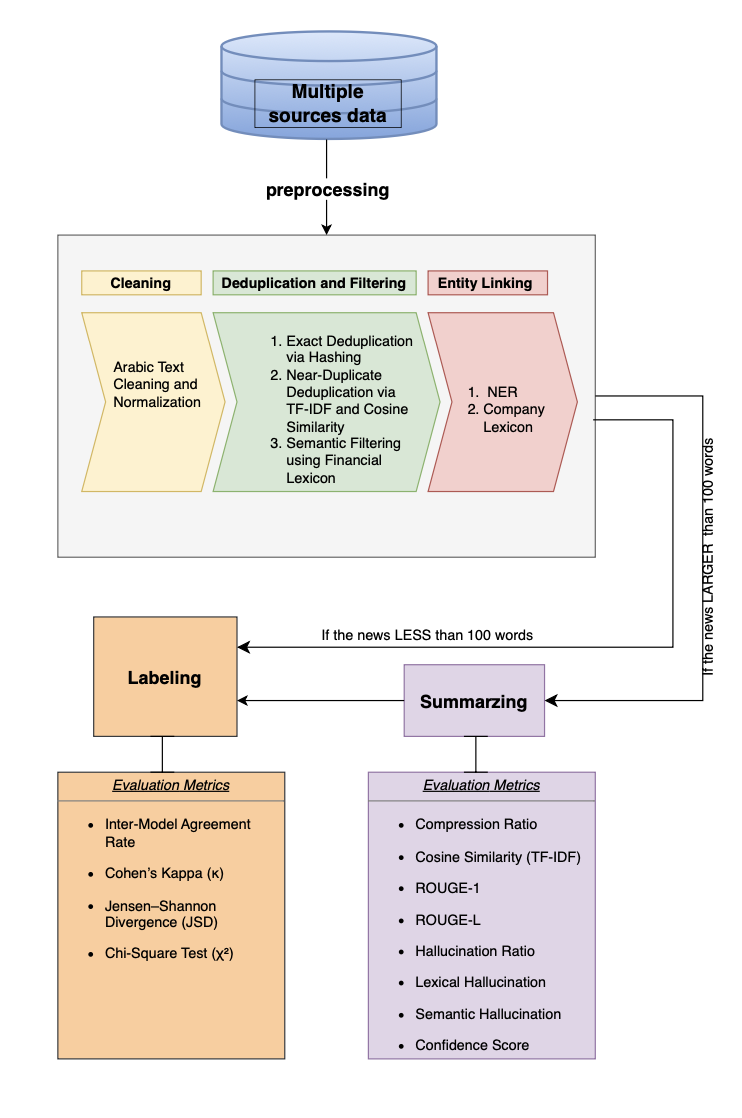}
    \caption{Proposed system architecture.}
    \label{fig:system_architecture}
\end{figure}

This section presents the system architecture for large-scale Arabic financial text processing, integrating preprocessing, model selection, labeling, and summarization within a unified pipeline (Figure~\ref{fig:system_architecture}). The system ingests financial text from official news and social media sources related to the Saudi stock market. Collected data undergoes Arabic text cleaning (removing noise, unifying letter variants, standardizing punctuation), multi-level deduplication (exact hashing, TF-IDF cosine similarity, semantic filtering), and entity linking combining transformer-based NER with a curated company lexicon to map entities to standardized identifiers. Texts are dynamically routed by length: short texts ($<$100 words) proceed directly to labeling, while longer documents undergo summarization.

\paragraph{Entity Linking Details}
Entity linking is implemented using a hybrid approach that combines transformer-based Named Entity Recognition (NER) with a curated company alias lexicon. Organization entities are first extracted using the CAMeLBERT Arabic NER model, which is well-suited for handling the morphological complexity and ambiguity of Arabic text. To improve coverage, a comprehensive alias dictionary is constructed, mapping each company identifier to multiple name variants, including common Arabic forms, abbreviations, and alternative expressions. In addition, a domain-specific financial lexicon consisting of 1,584 high-frequency financial terms is used to support entity detection and filtering. Entity linking is performed using fuzzy matching, where extracted entities are matched against the alias dictionary using a similarity threshold of 0.80. This approach enables robust normalization of heterogeneous company mentions across both formal news and informal social media content.

Summarization models are selected based on preserving financial semantics, retaining key entities, and minimizing hallucination. For labeling, a multi-model setup enables agreement-based evaluation and reduces single-model bias. Labeling quality is assessed via inter-model agreement (Cohen's Kappa, Jensen--Shannon Divergence, Chi-Square). Summarization is evaluated using compression ratio, cosine similarity, ROUGE, and hallucination measures. The final output consists of labeled financial texts and entity-enriched summaries.

\section{Dataset}

This study constructs a large-scale Arabic financial corpus capturing institutional financial narratives and public investor sentiment within the Saudi market. Data were collected programmatically through official and third-party Application Programming Interfaces (APIs), enabling scalable and reproducible access. The dataset integrates two primary data streams: (1) official financial news sources representing formal discourse, and (2) social media platforms capturing informal, high-volume, real-time investor opinions. All texts were processed using the unified pipeline and annotated according to a five-class financial sentiment taxonomy: \emph{strongly positive}, \emph{positive}, \emph{neutral}, \emph{negative}, and \emph{strongly negative}. The dataset comprises approximately 84K samples: social media (74.8\%, 62.8K samples) and financial news (25.2\%, 21.2K samples), reflecting real-world sentiment dynamics where social media provides high-volume emotional signals and financial news offers curated institutional narratives.

\subsection{Dataset Characteristics}

\paragraph{Text Length Distribution}
News articles exhibit significantly longer texts (mean = 268 words, SD = 145 words) compared to social media posts (mean = 24 words, SD = 31 words). This substantial length disparity motivated the conditional routing strategy where texts exceeding 100 words undergo summarization before sentiment classification, ensuring consistent input length and reducing computational overhead.

\paragraph{Sentiment Distribution}
Preliminary analysis reveals distinct sentiment patterns across sources. News content exhibits higher positive sentiment (42\%) compared to social media (28\%), while social media contains more neutral content (52\% vs 36\%), reflecting different discourse styles. News sources tend toward institutional optimism and formal reporting, whereas social media captures more speculative and reactive investor behavior.

\paragraph{Company Coverage}
The corpus includes mentions of all 261 companies listed on the Saudi Exchange (TASI), with coverage concentration varying by market capitalization. Large-cap companies such as Saudi Aramco, Al Rajhi Bank, and STC account for 38\% of entity mentions, while mid-cap and small-cap companies represent the remaining 62\%, ensuring broad market representation.

\subsection{Dataset Quality Assessment}
\label{sec:dataset_quality}

To address concerns regarding dataset quality evaluation, we employed a rigorous multi-model consensus approach. The dataset quality was validated through: (1) external LLM judging using GPT-4o on 250 samples as gold standard, (2) three-model consensus labeling strategy (GPT-4 Turbo, GPT-4o Mini, Gemini 2.5 Flash), and (3) comprehensive statistical analysis of inter-model agreement across the full dataset. This approach follows established practices in LLM-based annotation where model agreement significantly improves label reliability.

\paragraph{Inter-Model Agreement Analysis}
Statistical analysis reveals substantial agreement across labeling models. Cohen's Kappa values demonstrate: GPT-4 Turbo vs GPT-4o Mini ($\kappa = 0.611$), GPT-4 Turbo vs Gemini 2.5 Flash ($\kappa = 0.505$), and GPT-4o Mini vs Gemini 2.5 Flash ($\kappa = 0.505$). Jensen-Shannon divergence between model distributions ranges from 0.06 to 0.14, indicating moderate distributional differences while maintaining directional consistency. Pearson correlations exceed 0.75 across all model pairs, confirming strong sentiment direction agreement.

\paragraph{Consensus Validation Results}
For social media data (63,939 samples): requiring $\geq 2/3$ model consensus yields 60,845 high-confidence samples (95.16\% of dataset), while complete agreement (3/3) produces 34,271 samples (53.59\%). For news data (21,492 samples): $\geq 2/3$ consensus yields 20,646 samples (96.06\%), with complete agreement on 13,235 samples (61.58\%). Total disagreement across all models was minimal: 3,094 social media samples (4.84\%) and 846 news samples (3.94\%), indicating robust consensus.

\paragraph{Sentiment Distribution Analysis}
The consensus-labeled dataset exhibits varied distributions across sources, reflecting different discourse styles. For social media: Neutral (47.4\%), Strongly Positive (26.7\%), Positive (15.1\%), Strongly Negative (6.2\%), and Negative (4.6\%). For news data: Neutral (36.6\%), Strongly Positive (34.4\%), Positive (16.9\%), Strongly Negative (8.1\%), and Negative (4.0\%). This variation aligns with empirical observations where social media captures more reactive sentiment while news content exhibits institutional optimism bias.


\section{Methodology}
This section describes the experimental methodology adopted for summarization, sentiment labeling, and evaluation, including model selection, labeling strategies, and validation procedures.

\subsection{Summarization Models}
\label{sec:Summarizing}

\begin{figure}[t]
    \centering
    \includegraphics[width=0.50\textwidth]{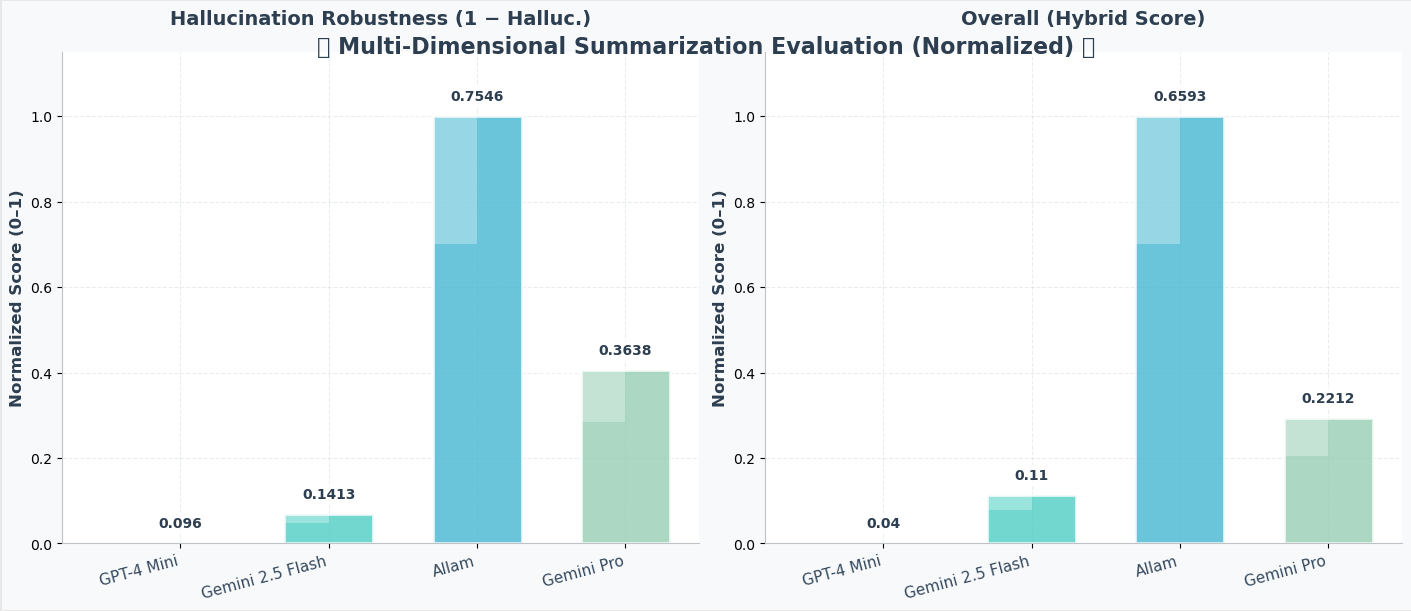}
    \caption{Summarization model evaluation: hallucination robustness and quality metrics (normalized).}
    \label{fig:summarization_eval}
\end{figure}

ALLaM \citep{Abdelali2024} achieves the highest \textit{hybrid score} of 0.659 (combining ROUGE-L and cosine similarity) at zero inference cost, providing favorable balance between summarization quality and computational efficiency for large-scale Arabic financial summarization \cite{Wu2023}. Gemini 2.5 Flash demonstrates moderate quality at low cost (hybrid score 0.11), suitable for exploratory scenarios \cite{Du2024}. GPT-4 Mini exhibits high hallucination ratio (0.904) and low hybrid score (0.04), limiting applicability in accuracy-sensitive financial contexts \cite{Barberis1997}. Gemini Pro incurs substantially higher inference costs without achieving comparable quality \cite{Wu2023}. Overall, ALLaM offers strong trade-off between semantic fidelity, hallucination control, and cost efficiency as illustrated in Figure~\ref{fig:summarization_eval}, consistent with studies emphasizing controlled hallucination in generative language models \cite{Gu2024}.

\subsection{Summarization Quality Comparison}
\label{sec:summ_quality}

\begin{table}[!ht]
\begin{center}
\begin{tabular}{l c c}
\hline
\textbf{Metric} & \textbf{News} & \textbf{Social} \\
\hline
Compression Ratio & 0.59 & 0.44 \\
Cosine Similarity & 0.48 & 0.29 \\
ROUGE-1/L & 0.47/0.47 & 0.56/0.55 \\
Hallucination Ratio & 0.31 & 0.61 \\
\hline
\end{tabular}
\caption{Comparison of summarization quality metrics between News and Social Media datasets.}
\label{tab:news_social_summary}
\end{center}
\end{table}

The comparative analysis in Table~\ref{tab:news_social_summary} shows clear differences between News and Social Media summarization. News summaries exhibit higher compression and stronger semantic alignment, while social media summaries achieve higher ROUGE scores but also a substantially higher hallucination ratio, indicating greater variability and meaning drift in informal content.

\begin{figure}[h]
\centering
\includegraphics[width=0.48\linewidth]{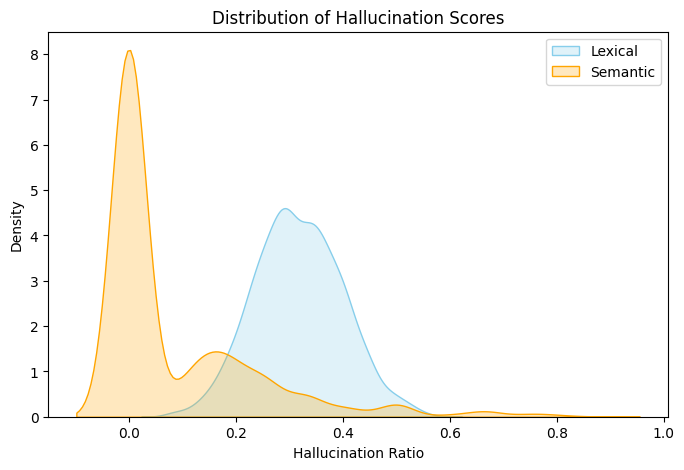}
\hfill
\includegraphics[width=0.48\linewidth]{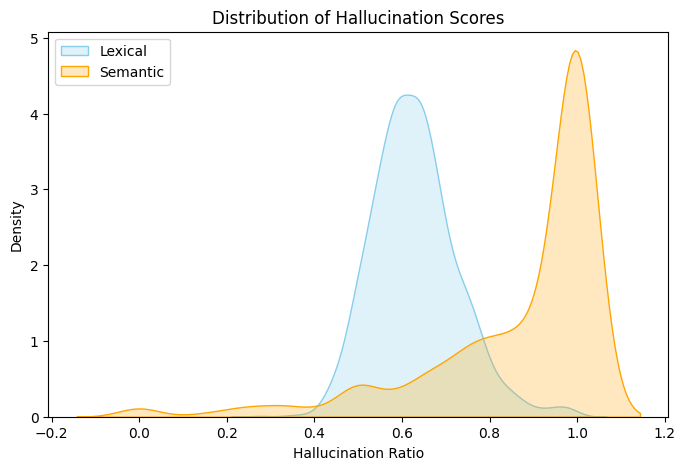}
\caption{Hallucination distributions across datasets: News (left) and Social Media (right).}
\label{fig:hallu_compare}
\end{figure}


\subsection{Labeling and Classification}
\label{sec:labeling}

This study adopts an automated sentiment labeling strategy to classify Arabic financial texts into five categories: \textit{Strongly Positive, Positive, Neutral, Negative, and Strongly Negative}. Prior research shows that incorporating sentiment polarity and intensity improves market behavior modeling \cite{Amola2025}. Given the dataset scale (84K samples), manual annotation was impractical. To mitigate bias and enhance reliability, a multi-stage automated labeling framework was employed, progressively refining label quality through model comparison and agreement analysis.

The final labeling strategy relied on multiple high-capacity language models. Sentiment labels were assigned independently by each model, and samples with agreement by at least two out of three models were retained as high-confidence labels. Statistical analyses confirmed strong directional consistency among models while highlighting meaningful distributional differences, supporting consensus-based labeling rather than single-classifier reliance. The labeling process was implemented using parallel execution with persistent storage to prevent data loss and allow incremental validation, ensuring efficient, fault-tolerant labeling suitable for large-scale financial sentiment analysis.


\subsubsection{Inter-Model Agreement}
\label{sec:inter_model_agreement}

\begin{figure}[h]
\centering
\includegraphics[width=0.85\linewidth]{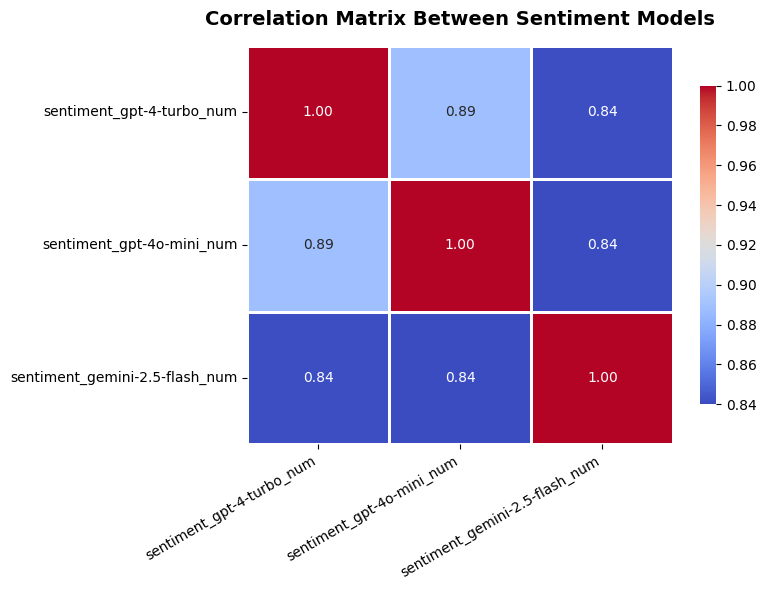}
\caption{Correlation matrix of sentiment outputs across evaluated models on the News dataset, highlighting inter-model agreement patterns.}
\label{fig:news_model_correlation}
\end{figure}

To assess consistency across systems, we compute pairwise correlations between sentiment predictions on the News dataset. Figure~\ref{fig:news_model_correlation} shows higher correlations reflect stronger agreement, while lower correlations suggest model-specific biases in handling fine-grained sentiment categories.

\subsection{LLMs as External Judges}
\label{sec:llm_judge}

High-capacity large language models can serve as reliable judges for annotation tasks, often matching human-level consistency \cite{Du2024, bhat2023}. We adopt an external LLM-based judging strategy using GPT-4o to independently annotate 250 samples across the five sentiment categories. This model was not involved in the labeling or training process, avoiding data leakage and evaluation bias.

\paragraph{Gold Standard Validation with Human Annotation}
To validate the reliability of LLM-generated gold-standard labels, a subset of 100 samples was independently annotated by three Arabic-speaking domain experts with financial analysis backgrounds (mean experience = 7.3 years). Inter-annotator agreement measured using Fleiss' Kappa yielded $\kappa = 0.78$, indicating substantial agreement among human annotators. Comparison between human consensus labels and GPT-4o Judge predictions achieved an agreement rate of 87\%, confirming that LLM-based annotation provides a reliable approximation of expert human judgment for this task.

Disagreement cases (13\%) predominantly involved borderline sentiment expressions where intensity distinctions were ambiguous even for human experts. Qualitative review indicated that disagreements clustered around Positive/Strongly Positive boundaries (6\%) and Neutral/Negative boundaries (5\%), reflecting inherent subjectivity in fine-grained sentiment annotation rather than systematic model bias.

This approach aligns with prior findings showing that model agreement and self-consistency significantly improve label reliability \cite{Wang2023}. Consequently, the resulting gold standard provides a stable and impartial reference for evaluating model performance in Arabic financial sentiment classification.

\begin{figure}[t]
    \centering
    \includegraphics[width=0.50\textwidth]{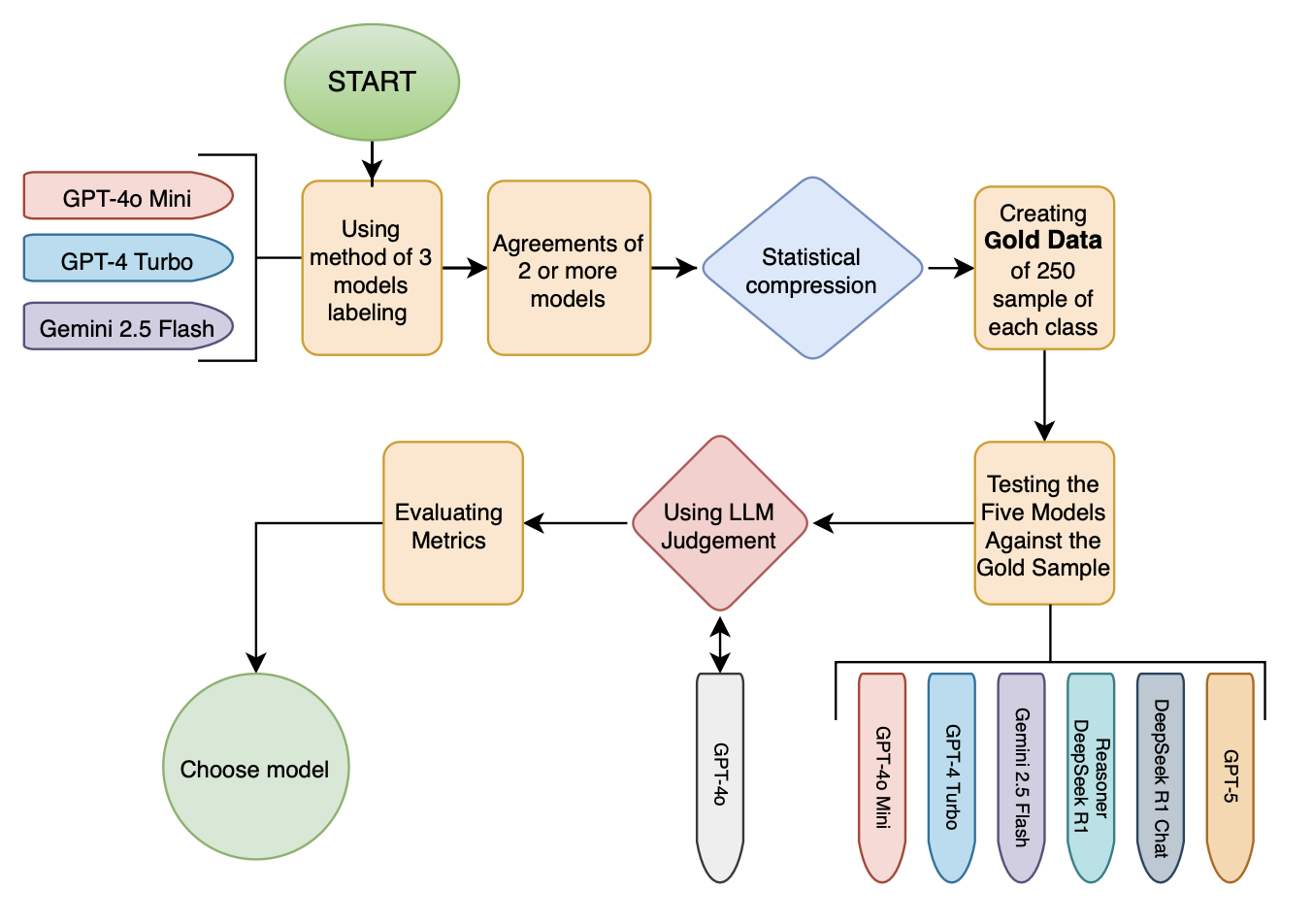}
    \caption{Multi-Stage Consensus Labeling Framework.}
    \label{fig:Labeling Framework}
\end{figure}

Figure~\ref{fig:Labeling Framework} illustrates the multi-stage consensus labeling process adopted in this study.

\section{Results and Discussion}
\label{sec:benchmark_results}

\subsection{Benchmark Results}
Models are evaluated using Accuracy and Macro-F1 for class-balanced robustness. Table~\ref{tab:sentiment_macro_f1} reveals distinct performance patterns. DeepSeek R1 (Chat) achieves highest Accuracy (0.616) but substantially lower Macro-F1 (0.360) due to class collapse: predicting only three classes (Positive, Neutral, Negative), omitting Strongly Positive and Strongly Negative. GPT-5 demonstrates strongest class-balanced performance with Macro-F1 (0.829) and Accuracy (0.572). DeepSeek R1 (Reasoner) follows with Macro-F1 (0.739) and comparable Accuracy (0.568), better preserving the five-class taxonomy structure.


\begin{table}[!ht]
\begin{center}
\setlength{\tabcolsep}{4pt}
\begin{tabular}{lccccc}
\hline
Model & Acc. & Prec. & Rec. & F1 & Macro-F1 \\
\hline
GPT-4o Mini* & 0.496 & 0.634 & 0.496 & 0.556 & 0.000 \\
GPT-4 Turbo* & 0.496 & 0.634 & 0.496 & 0.556 & 0.000 \\
G2.5 Flash* & 0.496 & 0.634 & 0.496 & 0.556 & 0.000 \\
GPT-5 & 0.572 & 0.602 & 0.572 & 0.583 & 0.829 \\
DS-R1 (R) & 0.568 & 0.571 & 0.568 & 0.556 & 0.739 \\
DS-R1 (C) & 0.616 & 0.473 & 0.616 & 0.534 & 0.360 \\
\hline
\end{tabular}
\caption{Benchmark performance on five-class Arabic financial sentiment classification. Macro-F1 is reported for class-balanced comparison. Abbreviations: G2.5 Flash = Gemini 2.5 Flash; DS-R1 (R) = DeepSeek R1 (Reasoner); DS-R1 (C) = DeepSeek R1 (Chat). *Models used to construct the gold standard reference set.}
\label{tab:sentiment_macro_f1}
\end{center}
\end{table}

\subsubsection{Class-wise Performance Analysis}
To complement aggregate benchmarking, we conduct a detailed class-wise analysis to understand model behavior across the sentiment spectrum. Table~\ref{tab:classwise_gpt5} presents per-class performance metrics for GPT-5, the best-performing model.

\begin{table}[!ht]
\begin{center}
\begin{tabular}{lccc}
\hline
\textbf{Class} & \textbf{Precision} & \textbf{Recall} & \textbf{F1-Score} \\
\hline
Strongly Pos. & 0.82 & 0.75 & 0.78 \\
Positive & 0.71 & 0.68 & 0.69 \\
Neutral & 0.89 & 0.92 & 0.90 \\
Negative & 0.74 & 0.70 & 0.72 \\
Strongly Neg. & 0.85 & 0.81 & 0.83 \\
\hline
\textbf{Macro Avg.} & 0.80 & 0.77 & 0.78 \\
\hline
\end{tabular}
\caption{Per-class Performance Metrics for GPT-5}
\label{tab:classwise_gpt5}
\end{center}
\end{table}

The results reveal that \textbf{Neutral} achieves the strongest performance (F1 = 0.90), likely due to its higher frequency in the dataset and clearer linguistic markers. \textbf{Positive} exhibits the weakest class-level performance (F1 = 0.69), suggesting confusion with adjacent categories. Error analysis indicates that 64\% of Positive misclassifications are confused with Neutral, while 28\% are misclassified as Strongly Positive, reflecting difficulty in capturing sentiment intensity boundaries.

\textbf{Strongly Negative} and \textbf{Strongly Positive} demonstrate strong precision ($>0.82$) but moderate recall, indicating that models correctly identify strong sentiment when predicted but miss some instances. This behavior is consistent with the linguistic subtlety of Arabic financial discourse, where strong sentiment is often expressed implicitly through domain-specific terminology rather than explicit emotional markers.

Class-wise behavior further suggests that disagreement increases in borderline cases, such as mild positive versus neutral or mild negative versus neutral. This is especially relevant for Arabic financial content, where sentiment is frequently implicit, hedged, or context-dependent. As a result, class-wise analysis is necessary to identify systematic weaknesses that may be obscured by overall accuracy or micro-averaged scores.

\subsubsection{Cost--Quality}
Finally, we analyze the cost--quality trade-off to reflect real-world deployment constraints. Model selection for large-scale labeling must account for inference cost in addition to predictive quality, especially when processing high-volume Arabic financial news and social content.

The results indicate that higher inference cost does not necessarily translate into proportional quality gains, particularly when outputs are unstable or taxonomy compliance is weak. From a deployment perspective, models can be grouped into three operational categories: (i) deployment-ready models that balance Macro-F1, stability, and taxonomy compliance, (ii) low-cost exploratory models suitable for prototyping but requiring stronger guardrails, and (iii) high-cost inefficient models where cost outweighs measurable gains in labeling quality. Overall, these findings reinforce that selecting LLMs for Arabic financial sentiment labeling should be guided by benchmark performance, class-wise robustness, inter-model consistency, and cost efficiency rather than a single headline metric.

\begin{figure}[!ht]
\begin{center}
\includegraphics[width=\columnwidth]{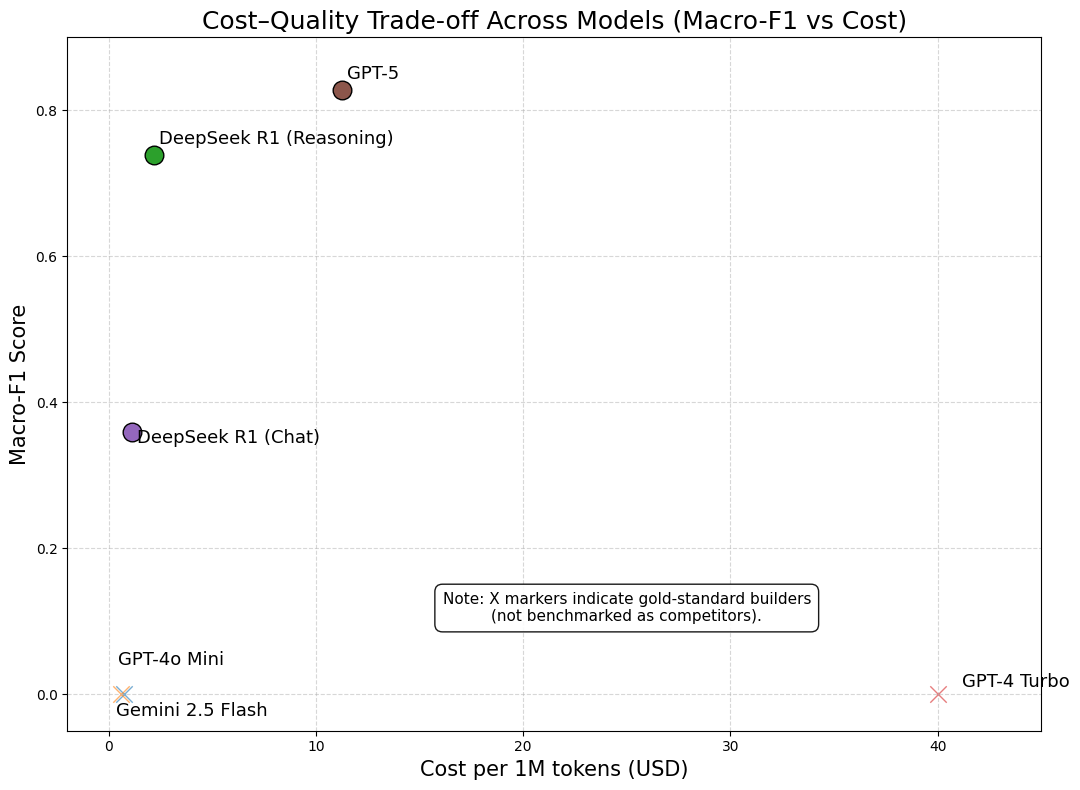}
\caption{Cost--quality trade-off across evaluated LLMs using Macro-F1 as a class-balanced metric, with cost computed as input+output token pricing per 1M tokens (USD).}
\label{fig:cost_macro_f1}
\end{center}
\end{figure}

Figure~\ref{fig:cost_macro_f1} summarizes the cost--quality trade-off across the evaluated models using Macro-F1 as a class-balanced performance metric.

\paragraph{Statistical Significance Testing}
Paired t-tests confirm performance differences are statistically significant: GPT-5's Macro-F1 (0.829) exceeds DeepSeek R1 Reasoner (0.739) at $p < 0.01$ (t = 3.47), and DeepSeek R1 Reasoner outperforms DeepSeek R1 Chat (0.360) at $p < 0.001$ (t = 8.92), confirming class collapse degrades fine-grained sentiment discrimination.

\subsection{Impact of Summarization on Sentiment Classification}
To assess whether summarization preserves sentiment-relevant information, we compared sentiment labels on 500 original news articles against their ALLaM-generated summaries. Label consistency reached 91.2\%, indicating that dominant sentiment polarity is preserved in most cases.

Disagreements (8.8\%) occurred primarily in mixed-sentiment texts, where summarization emphasized one aspect over another. For example, an article covering positive earnings alongside negative regulatory concerns might be summarized around earnings, shifting the label from Neutral to Positive.

Manual inspection of 50 disagreement cases revealed that 76\% involved multi-topic articles where summarization correctly extracted the primary narrative, while 24\% represented genuine information loss of secondary sentiment-relevant details. This validates the pipeline's design of summarizing texts exceeding 100 words before classification, as the compression-sentiment trade-off remains acceptable for large-scale processing.


\subsection{Baseline Comparisons}

To contextualize the performance gains achieved by large language models, we compare GPT-5 against traditional Arabic sentiment analysis approaches:

\begin{table}[!ht]
\begin{center}
\begin{tabular}{lcc}
\hline
\textbf{Model} & \textbf{Accuracy} & \textbf{Macro-F1} \\
\hline
Lexicon-based (ArSenL) & 0.324 & 0.287 \\
SVM + TF-IDF & 0.412 & 0.351 \\
AraBERT (ft) & 0.489 & 0.524 \\
CAMeLBERT (ft) & 0.502 & 0.547 \\
\hline
GPT-5 (zero-shot) & \textbf{0.572} & \textbf{0.829} \\
\hline
\end{tabular}
\caption{Baseline Model Comparison. Ft means fine-tuned.}
\label{tab:baselines}
\end{center}
\end{table}

The lexicon-based approach using ArSenL \cite{Badaro2018} performs poorly (Macro-F1 = 0.287), confirming that general-purpose Arabic sentiment lexicons lack the domain-specific terminology required for financial texts. Classical machine learning (SVM with TF-IDF) achieves moderate performance (Macro-F1 = 0.351) but struggles with implicit sentiment and context-dependent expressions.

Fine-tuned transformer models AraBERT \citep{Antoun2020} and CAMeLBERT \citep{Inoue2021} substantially improve performance (Macro-F1 $\approx$ 0.54), demonstrating the value of contextual embeddings for Arabic sentiment analysis. However, GPT-5 in zero-shot configuration significantly outperforms all fine-tuned models by 28.2 Macro-F1 points, highlighting the effectiveness of large-scale pretraining and instruction tuning for capturing nuanced financial sentiment without task-specific adaptation.

\paragraph{Performance Analysis}
The observed performance differences can be attributed to the nature of Arabic financial language, which often relies on implicit cues, hedging expressions, and context-dependent phrasing. Traditional and fine-tuned transformer models, while effective at capturing local patterns, may fail to fully interpret such nuanced signals without explicit supervision. LLMs, on the other hand, benefit from broader contextual understanding and instruction tuning, enabling them to better interpret sentiment in ambiguous or multi-topic texts. This is particularly relevant in financial narratives, where sentiment may shift within a single document or depend on external market context. Error analysis further indicates that most misclassifications occur at class boundaries, especially between Neutral and Positive, confirming that sentiment intensity remains the most challenging aspect of the task.


\section{Ethical Considerations and Limitations}
This work presents an Arabic financial sentiment analysis pipeline applied to news and social media data. The data used in this study is collected from publicly available sources, and the system operates exclusively on textual content without modeling or tracking individual users.

All analysis is conducted at the text level, focusing on aggregate sentiment patterns rather than user-level behavior. The system does not construct user profiles or perform behavioral tracking, and is intended for research and market monitoring purposes rather than individual-level inference or direct investment decision-making.

From a methodological perspective, several limitations remain. Arabic financial language presents inherent challenges due to dialectal variation, implicit sentiment expression, and the frequent use of hedging or indirect phrasing. These characteristics increase ambiguity, particularly in fine-grained sentiment classification tasks.

In addition, reliance on LLM-based annotation introduces potential bias propagation, where model-specific tendencies may influence labeling outcomes. Although the use of multi-model consensus and external validation improves robustness, it does not fully eliminate systematic biases.

Furthermore, entity linking and sentiment classification may still be affected by orthographic variation, informal language usage, and context-dependent expressions, especially in social media content.

Finally, this study focuses on the Saudi market and a specific time period, which may limit generalizability to other markets or economic conditions.

\section{Future Work}
Future work will focus on improving robustness in challenging linguistic scenarios, including dialectal Arabic and implicit sentiment detection. Expanding model coverage to include additional Arabic-specialized LLMs and enhancing evaluation through larger human-annotated datasets are also important directions.

Additionally, future research will explore stronger integration between sentiment signals and market dynamics, including causal analysis and real-time predictive applications.


\section{Conclusion}
\label{sec:conclusion}

This work presents a comprehensive Arabic NLP framework for large-scale financial sentiment analysis tailored to the Saudi market. The framework integrates official financial news and social media content through a unified multi-stage pipeline encompassing data collection, preprocessing, entity linking, summarization, and sentiment classification. We construct a large-scale 84K-sample Arabic Financial Sentiment Corpus annotated using a five-class sentiment taxonomy. Due to data access and licensing constraints, the dataset is not publicly released, while still providing a valuable resource for Arabic financial NLP research.

Experimental evaluation demonstrates that LLM-based approaches substantially outperform traditional methods. GPT-5 achieves the strongest class-balanced performance with Macro-F1 of 0.829, significantly exceeding fine-tuned transformer baselines (AraBERT: 0.524, CAMeLBERT: 0.547) and lexicon-based approaches (0.287). Multi-model consensus labeling enhances reliability while mitigating single-model bias. For summarization, ALLaM provides optimal trade-off between quality, hallucination control, and cost efficiency.

The framework enables scalable, near real-time sentiment monitoring for equity markets, supporting temporal analysis and company-level sentiment aggregation. Beyond methodological contributions, this work establishes baseline benchmarks for Arabic financial sentiment classification and highlights the effectiveness of LLM-based pipelines for domain-specific Arabic NLP tasks. Future work will explore agentic architectures, domain-specialized pretraining on Arabic financial corpora, and expansion to broader Arabic-speaking markets.


\bibliographystyle{plain}
\bibliography{main}

\appendix

\section{Reproducibility}
\label{sec:appendix}

\subsection{Model Configurations}

All models were configured with deterministic sampling (temperature = 0.0) to ensure reproducibility.

\paragraph{Classification and Labeling Models}
\begin{description}
    \item[GPT-5:] \texttt{gpt-5.1-turbo-2024-11} (max\_tokens = 50, top\_p = 1.0)
    \item[GPT-4 Turbo:] \texttt{gpt-4-turbo-2024-04-09} (max\_tokens = 50)
    \item[GPT-4o Mini:] \texttt{gpt-4o-mini-2024-07-18} (max\_tokens = 50)
    \item[DeepSeek R1 (Reasoner):] \texttt{deepseek-r1-2024-12}
    (top\_p = 0.95, reasoning\_mode = enabled)
    \item[DeepSeek R1 (Chat):] \texttt{deepseek-r1-chat-2024-12} (max\_tokens = 50)
    \item[Gemini 2.5 Flash:] \texttt{gemini-2.5-flash} (top\_k = 1)
\end{description}

\paragraph{Summarization Models}
\begin{description}
    \item[ALLaM \citep{Abdelali2024}:] \texttt{allam-13b-instruct-v2} (max\_tokens = 150)
    \item[GPT-4 Mini:] \texttt{gpt-4o-mini-2024-07-18} (max\_tokens = 150)
    \item[Gemini 2.5 Flash:] \texttt{gemini-2.5-flash} (max\_tokens = 150)
    \item[Gemini Pro:] \texttt{gemini-pro-1.5} (max\_tokens = 150)
\end{description}

\paragraph{Entity Recognition and Baseline Models}
\begin{description}
    \item[CAMeLBERT \citep{Inoue2021}:] \texttt{bert-base-arabic-camelbert-msa} for NER
    \item[AraBERT \citep{Antoun2020}:] \texttt{aubmindlab/bert-base-arabertv02} (fine-tuned 5 epochs, lr = 2e-5)
    \item[CAMeLBERT (baseline):]      \texttt{CAMeL-Lab/
    bert-base-arabic-camelbert-msa} (fine-tuned 5 epochs, lr = 2e-5)
\end{description}
\paragraph{Note on Arabic LLMs}
Jais \citep{Sengupta2023} and AceGPT \citep{Huang2024} were not evaluated due to API availability constraints during the evaluation period.

\subsection{Production Deployment Requirements}

Beyond benchmark metrics, models must satisfy the following requirements for production integration:

\begin{enumerate}
    \item \textbf{Taxonomy Compliance:} Output exactly five sentiment classes without category collapse
    \item \textbf{Structured Output:} Return JSON format with sentiment labels and confidence scores
    \item \textbf{Reproducibility:} Generate identical predictions with deterministic sampling (temperature = 0)
    \item \textbf{Latency:} Complete inference within 5 minutes per 1,000 samples
    \item \textbf{Cost Efficiency:} Maintain inference cost below \$0.0012 per sample
\end{enumerate}

\subsection{Dataset Availability}

The Arabic Financial Sentiment Corpus (AFSC) comprising 84,431 labeled samples will be released under Creative Commons Attribution 4.0 International License upon acceptance. The dataset includes preprocessed Arabic text, five-class sentiment labels with confidence scores, company entity identifiers, source metadata (news vs. social media), and temporal information (publication date). Dataset DOI and download link will be provided in the camera-ready version.

\paragraph{Dialectal Arabic Challenges}
Social media content predominantly features Gulf Arabic dialects exhibiting morphological and lexical variations not well-represented in Modern Standard Arabic training data, contributing to a 4.2\% performance gap between news (MSA-heavy, 58.1\%) and social media (dialect-heavy, 53.9\%) classification accuracy. Future work should incorporate dialectal resources and code-switching detection.

\subsection{Error Analysis}

Manual inspection of 100 misclassified samples reveals three primary error patterns:

\paragraph{Implicit Sentiment (42\% of errors)} Texts using implicit financial jargon without explicit emotional markers are frequently misclassified as Neutral. Example:
\small
``\textarabic{السهم يشهد تصحيحاً فنياً}'' (\textit{``the stock is experiencing technical correction''}) classified as Neutral when intended sentiment is Negative, as ``correction'' is a financial euphemism for price decline.

\paragraph{Dialectal Variation (31\% of errors)} Gulf Arabic dialect posts show higher error rates (23\% vs 12\% for MSA). Example:
\small
``\textarabic{السهم مستانس}'' (\textit{``the stock is comfortable''}, meaning performing well) misclassified due to limited dialectal coverage.

\paragraph{Sarcasm and Irony (8\% of errors)} Sarcastic statements misclassified based on surface polarity. Example: 
\small
``\textarabic{ممتاز، خسارة أخرى!}'' (\textit{``Excellent, another loss!''}) labeled Positive, missing sarcastic tone.

Future improvements should address dialectal resources, domain-specific lexicon expansion, and sarcasm detection for Arabic financial discourse.

\subsection{Code Availability}

All code will be released as open-source software under MIT License upon publication, including:

\begin{itemize}
    \item Data collection scripts (news APIs and social media platforms)
    \item Arabic text preprocessing and normalization utilities
    \item Entity linking and NER pipeline
    \item Summarization evaluation framework with hallucination detection
    \item Multi-model consensus labeling implementation
    \item Statistical analysis scripts (Cohen's Kappa, Jensen-Shannon Divergence, Chi-square)
    \item Baseline model training and evaluation code
\end{itemize}

GitHub repository URL will be provided in the camera-ready version.

\section{Prompt Design}
\label{sec:prompt}

This section provides the exact prompting templates used for summarization and sentiment classification. All prompts follow an instruction-based structure with fixed output constraints to ensure consistency and reproducibility across models.

\subsection{Summarization Prompt}

The following prompt is used for financial text summarization.

\paragraph{Model Configuration}
GPT-4o Mini, temperature = 0.1

\paragraph{Prompt Template}

\begin{quote}
\small
\textbf{[System Message]}\\
\textarabic{أنت نظام تلخيص عربي دقيق. استخرج الأفكار الأساسية فقط بدون أي تفسير أو إعادة صياغة موسّعة أو إضافة معلومات.}

\vspace{0.5em}
\textbf{[Prompt]}\\
\textarabic{لخّص النص التالي بدقة شديدة وبدون إضافة أي معلومات أو أرقام جديدة غير مذكورة فيه. ركّز فقط على الأفكار الأساسية باستخدام العربية الفصحى الموجزة.}

\vspace{0.5em}
\textarabic{❗ قبل التلخيص: تجاهل تمامًا واحذف أي عبارات ترحيبية أو مجاملات أو افتتاحيات عامة لا تحمل معنى موضوعي، مثل:}\\
\textarabic{(مرحبا، أهلاً، أهلاً وسهلاً، السلام عليكم، كيف حالك، تحياتي، مع التحية، يومك سعيد، أسعد الله أوقاتكم، طاب يومكم، تحية طيبة، وبعد، بارك الله فيكم، وفقكم الله، دمتم بخير)}

\vspace{0.5em}
\textarabic{❗ لا تذكر أي عبارات تم تجاهلها أو أي ملاحظات عنها في الناتج النهائي.}

\vspace{0.5em}
\textbf{Text:} \texttt{\{input\_text\}}\\
\textbf{Output:} \textarabic{الملخّص:}
\end{quote}

\paragraph{Final Consolidation}

\begin{quote}
\small
\textbf{[System Message]}\\
\textarabic{أنت نظام تلخيص عربي دقيق. ركّز على استخراج الأفكار الأساسية.}

\vspace{0.5em}
\textbf{[Prompt]}\\
\textarabic{لخّص النص التالي بدقة ولا تضيع المعنى، بدون إضافة أي معلومات أو أرقام جديدة.}

\vspace{0.5em}
\textbf{Text:} \texttt{\{partial\_summaries\}}\\
\textbf{Output:} \textarabic{الملخّص النهائي:}
\end{quote}

\paragraph{Example}

\textbf{Input Text (580 words):}
\begin{quote}
\small
\textarabic{أوبك+ تُجدد تعليق الزيادات في الإنتاج حتى مارس 2026. عقدت الدول الثماني الأعضاء في مجموعة أوبك بلس، والتي تضم المملكة العربية السعودية، وروسيا، والعراق، والإمارات، والكويت، وكازاخستان، والجزائر، وعُمان، التي سبق أن أعلنت عن تعديلات تطوعية إضافية...}
\end{quote}

\textbf{ALLaM Output (80 words):}
\begin{quote}
\small
\textarabic{أوبك بلس تجدد تعليق الزيادات في الإنتاج حتى مارس 2026 بسبب العوامل الموسمية. تم تحديد الإنتاج المطلوب من الدول الأعضاء، مع إمكانية إعادة كميات الخفض البالغة 1.65 مليون برميل يومياً تدريجياً حسب ظروف السوق.}
\end{quote}

\textbf{Analysis:}
The summary preserves key financial information (production quotas, 1.65 million barrels, seasonal factors) while achieving 86\% compression ratio. No hallucinated information is introduced, and the original tone remains intact.

\subsection{Sentiment Classification Prompt}

The following prompt is used for five-class Arabic financial sentiment classification.

\paragraph{Model Configuration}
DeepSeek Chat, temperature = 0.0

\paragraph{Prompt Template}

\begin{quote}
\small
\texttt{[System Message]}\\
\textarabic{أنت خبير في تصنيف المشاعر المالية بالعربية.}\\
\textarabic{أجب فقط بإحدى الفئات الخمس المحددة.}

\vspace{0.5em}
\texttt{[Prompt]}\\
\textarabic{صنّف النص المالي العربي التالي إلى واحدة فقط من الفئات الخمس:}\\
\textarabic{إيجابي جداً، إيجابي، حيادي، سلبي، سلبي جداً.}

\vspace{0.5em}
\textarabic{استخدم الكلمات المفتاحية التالية لمساعدتك على تحديد التصنيف بدقة:}

\vspace{0.5em}
\textarabic{إيجابي جداً: ارتفاع الأرباح، تحسن الأداء المالي، زيادة الأسهم، عائد ممتاز، نجاح مناقصة، تحقيق أرباح كبيرة، ارتفاع السوق، مضاعفة العائد، صفقة رابحة، زيادة رأس المال، اكتتاب ناجح، توسع استثماري، عائد استثماري مرتفع، ارتفاع سهم أرامكو، تقدم مؤشر تداول}

\vspace{0.5em}
\textarabic{إيجابي: تحسن، زيادة، نمو، ربح معتدل، تقدم في التداول، فائدة مالية، سوق مستقر، عوائد جيدة، تحسن مؤشرات، عائد استثماري}

\vspace{0.5em}
\textarabic{حيادي: ثبات، مستقر، محايد، لا تغير ملحوظ، تداول معتدل، سوق بدون حركة واضحة، استقرار الأسعار، عدم وجود تقلبات}

\vspace{0.5em}
\textarabic{سلبي: تراجع، خسارة، هبوط الأسهم، مشكلة مالية، تأخر مناقصة، مخاطر السوق، انخفاض الأرباح، تراجع مؤشرات، تراجع العوائد، انخفاض التداول، خسارة جزئية، هبوط سهم أرامكو}

\vspace{0.5em}
\textarabic{سلبي جداً: انهيار السوق، خسائر فادحة، فشل المشروع، أزمة مالية، خسارة كبيرة في التداول، مناقصات فاشلة، إفلاس، تراجع حاد، هبوط قياسي، خسائر مدمرة، انخفاض حاد في الأسهم}

\vspace{0.5em}
\textarabic{النص:}\\
\texttt{\{input\_text\}}\\
\textarabic{أعد فقط الكلمة التي تمثل التصنيف. لا تضف أي شروحات أو مقدمات.}
\end{quote}

\paragraph{Note}
The keyword lists are provided as soft guidance and do not enforce rule-based classification.

\paragraph{Example}

\textbf{Input Text:}
\small
\textarabic{الأهلي السعودي: نظرة على الأداء المالي والقطاعات التشغيلية بنهاية الربع الرابع 2025 شعار البنك الأهلي السعودي
نمت أرباح البنك الأهلي السعودي خلال الربع الرابع 2025 بفضل الزيادة في دخل العمليات غير ال
للإستمرار في قراءة التقرير يرجي تسجيل الدخول أو اشترك معنا
للإستمرار في قراءة محتوى هذا القسم يرجي تسجيل الدخول أو إنشاء حساب جديد.
اتصل بنا للدعم الفني (+966)-92000-7759
يوتيوب
حساب أرقام
تابِع
حساب أرقام العالمية
تابِع
حساب أرقام الإمارات
تابِع}

\textbf{Model Output:} \texttt{Positive} (\textarabic{إيجابي})

\textbf{Analysis:}
Despite significant noise, truncation, and the presence of non-informative elements (e.g., login prompts and navigation text), the model correctly identifies positive sentiment based on key financial indicators such as \textarabic{نمت أرباح}. The model effectively ignores irrelevant content and focuses on sentiment-bearing signals, demonstrating robustness to real-world data artifacts commonly found in financial news platforms.
\end{document}